\documentclass[letterpaper, 10 pt, conference]{ieeeconf}  
\makeatletter
\def\@citex[#1]#2{\leavevmode
	\let\@citea\@empty
	\@cite{\@for\@citeb:=#2\do
		{\@citea\def\@citea{,\penalty\@m\ }%
			\edef\@citeb{\expandafter\@firstofone\@citeb\@empty}%
			\if@filesw\immediate\write\@auxout{\string\citation{\@citeb}}\fi
			\@ifundefined{b@\@citeb}{\hbox{\reset@font\bfseries ?}%
				\G@refundefinedtrue
				\@latex@warning
				{Citation `\@citeb' on page \thepage \space undefined}}%
			{\@cite@ofmt{\csname b@\@citeb\endcsname}}}}{#1}}
\makeatother

\usepackage{graphicx}
\usepackage{amsmath}

\newcommand\T{\rule{0pt}{2.6ex}}       
\newcommand\B{\rule[-1.2ex]{0pt}{0pt}} 

\usepackage{tabularx}
\usepackage{multirow, multicol}
\usepackage[table,xcdraw]{xcolor}
\usepackage{colortbl}

\IEEEoverridecommandlockouts                              

\makeatletter
\newcommand*\titleheader[1]{\gdef\@titleheader{#1}}
\AtBeginDocument{\let \st@red@title \@title \def \@title{\bgroup \normalfont \footnotesize \flushleft \@titleheader \par \egroup \vskip0.5em \st@red@title}}
\makeatother

\title{\LARGE \bf
Multi-Task Recurrent Neural Network for Surgical Gesture Recognition and Progress Prediction}

\titleheader{
Multi-Task Recurrent Neural Network for Surgical Gesture Recognition and Progress Prediction. Beatrice van Amsterdam, Matthew J Clarkson, Danail Stoyanov, In 2020 IEEE International Conference on Robotics and Automation (ICRA).\\
\copyright2020 IEEE. Personal use of this material is permitted. Permission from IEEE must be obtained for all other uses, in any current or future media, including reprinting/republishing this material for advertising or promotional purposes, creating new collective works, for resale or redistribution to servers or lists, or reuse of any copyrighted component of this work in other works.}

\author{Beatrice van Amsterdam$^{1}$, Matthew J. Clarkson$^{1}$, Danail Stoyanov$^{1}$
\thanks{This work was supported by the Wellcome/EPSRC Centre for Interventional and Surgical Sciences (WEISS) at UCL (203145Z/16/Z), EPSRC (EP/P027938/1, EP/R004080/1,NS/A000027/1), the H2020 FET (GA 863146) and Wellcome [WT101957]. Danail Stoyanov is supported by a Royal Academy of Engineering Chair in Emerging Technologies (CiET1819/2/36) and an EPSRC Early Career Research Fellowship (EP/P012841/1).}
\thanks{$^{1}$B. van Amsterdam, M. J. Clarkson and D. Stoyanov are with the Wellcome/EPSRC Centre for Interventional and Surgical Sciences (WEISS), University College London, London, United Kingdom.
{\tt\small beatrice.amsterdam.18@ucl.ac.uk}}%
}

\begin{document}

\maketitle
\thispagestyle{empty}
\pagestyle{empty}

\begin{abstract}

Surgical gesture recognition is important for surgical data science and computer-aided intervention. Even with robotic kinematic information, automatically segmenting surgical steps presents numerous challenges because surgical demonstrations are characterized by high variability in style, duration and order of actions. In order to extract discriminative features from the kinematic signals and boost recognition accuracy, we propose a multi-task recurrent neural network for simultaneous recognition of surgical gestures and estimation of a novel formulation of surgical task progress. 
To show the effectiveness of the presented approach, we evaluate its application on the JIGSAWS dataset, that is currently the only publicly available dataset for surgical gesture recognition featuring robot kinematic data. We demonstrate that recognition performance improves in multi-task frameworks with progress estimation without any additional manual labelling and training.

\end{abstract}


\section{Introduction}

Automated surgical gesture recognition aims at automatically identifying meaningful action units within surgical tasks that constitute a surgical intervention. 
The process forms a fundamental step in the development of systems for surgical data science \cite{MaierHein2017}, objective skill evaluation \cite{Vedula2016, Reiley2009} and surgical automation \cite{Reiley2010, Murali2015, Nagy2019}. The problem is however challenging because surgical gestures have high degree of variability due to multiple parameters in the operating surgeon's style and the patients' anatomy which alters the duration, kinematics and order of actions among different demonstrations \cite{Cao1996}. 

Much research in the field, however, is based on the premise that many surgical tasks have well-defined structure and use specific action patterns to progress towards a surgical goal. 
Gesture flow has then been described through task-specific probabilistic grammars \cite{Ahmidi2017}, which have been modelled with powerful statistical tools such as graphical models \cite{Varadarajan2009, Tao2012} and neural networks \cite{Lea2016tcnECCV, DiPietro2016}. 
This work investigates if the recognition performance improves when the progress of the surgical task is modelled explicitly and learnt jointly with the action sequence, resulting in a more discriminative feature extraction process.

The effectiveness of multi-task learning \cite{Ruder2017} and surgical progress modelling has been demonstrated in previous work focused on surgical workflow analysis \cite{Yengera2018,Li2017}, where the aim is to recognise surgical phases representing high-level surgical states. We adopt this approach with high-granularity gesture sequences and design a multi-task recurrent neural network for simultaneous gesture recognition and progress estimation. Differently from previous work, however, the task progress is based on the underlying action sequence rather than on time. We hypothesize that action-based progress estimation could help to learn action sequentiality despite duration variability and the presence of adjustment gestures and spurious motions, and thus reduce out-of-order predictions and over-segmentation errors. 
We also analyse different progress estimation strategies and highlight correlations between gesture and progress predictions.

We validate our algorithm on the kinematic data of the JIGSAWS dataset \cite{Gao2014}, featuring demonstrations of elementary surgical tasks collected from eight surgeons with different skill level using the da Vinci Surgical System (dVSS, Intuitive Surgical Inc.) \cite{Guthart2000}. Our experiments show that gesture recognition performance improves in multi-task frameworks with progress estimation at no additional cost, as the progress labels can be generated automatically from the data and available action labels.

\subsection{Related Work}

\begin{figure*}[t!]
	\vspace{0.05 cm}
	\centering
	\includegraphics[width=\textwidth]{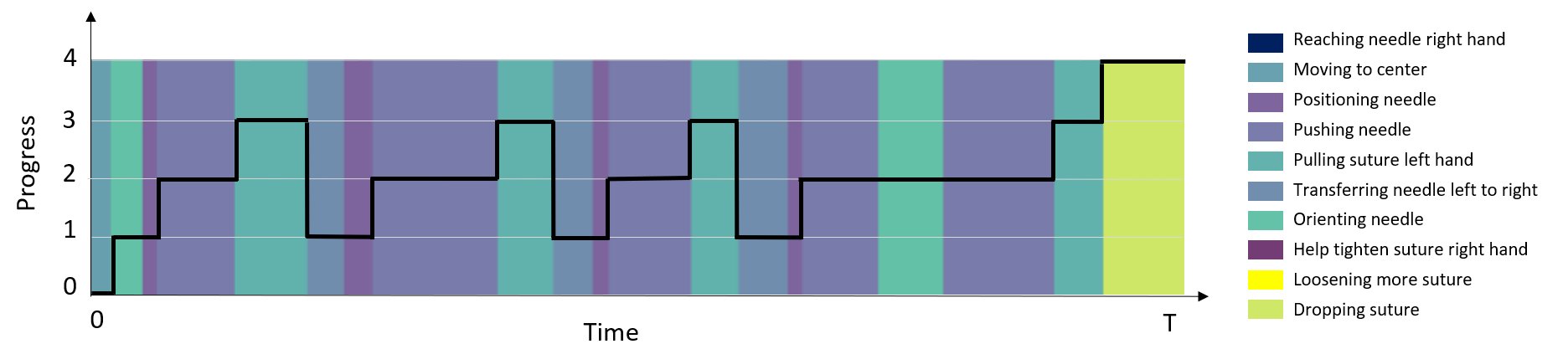}
	\caption{Our definition of progress is dictated by the underlying action sequence. We identified five gestures that represent essential progressive stages in any complete suturing demonstration. The other classes represent adjustment gestures that serve to prepare or help to complete the execution of the essential gestures.}
	\label{fig:progLabels}
\end{figure*}

Gesture recognition from robot kinematics has been tackled through probabilistic graphical models such as Hidden Markov Models (HMMs) \cite{Varadarajan2009, Tao2012, Sefati2015} and Conditional Random Fields (CRFs) \cite{Tao2013, Lea2015semantic, Mavroudi2018}. These however rely on frame-to-frame and segment-to-segment transitions only, 
ignoring long-range temporal dependencies in the surgical demonstrations. 
Deep learning techniques have been recently used to capture complex, long-distance patterns through hierarchies of temporal convolutional filters \cite{Lea2016tcnECCV, Lea2017tcnCVPR}, LSTM networks \cite{DiPietro2016} or deep Reinforcement Learning (RL) \cite{Liu2018}. 
Besides, unsupervised \cite{Krishnan2015, Fard2017} and weakly-supervised \cite{vanAmsterdam2019} recognition have been shown through clustering, which reduces the dependency on annotations but at the expense of performance. 

Surgical video rather than kinematics also embeds gesture information which can be extracted with spatio-temporal CNNs \cite{Lea2016}, 3D CNNs \cite{Funke2019}, multi-scale temporal convolutions \cite{Lei2018, Wang2019} or hybrid encoder-decoder networks with temporal-convolutional filters for local motion modelling and bidirectional LSTM for long-range dependency memorization \cite{Ding2017}. 
 
Finally, a number of studies have approached surgical workflow analysis through multi-task learning. 
Examples include systems for joint task and gesture classification \cite{Sarikaya2018}, and models for joint phase recognition and tool detection \cite{Twinanda2017} or progress estimation \cite{Li2017}. Phase recognition networks have also been pre-trained on auxiliary tasks such as prediction of the  Remaining Surgery Duration (RSD) \cite{Yengera2018} or estimation of the frame temporal order \cite{Bodenstedt2017}, aiming to improve understanding of the temporal progression of the surgical workflow.
Such approaches show that multi-task learning and progress modelling are beneficial for surgical workflow understanding and could support fine-grained analysis that requires discriminative feature extraction.

\section{Methods}

\subsection{Dataset}

We trained our network on the 39 suturing demonstrations of the JIGSAWS dataset, using the kinematic data (end-effector position, velocity, gripper angle) recorded at 30 Hz from the two Patient Side Manipulators (PSMs) of the dVSS. The trajectories were first smoothed with a low-pass filter with cut-off frequency $f_c=1.5$ Hz against measurement noise \cite{Despinoy2016}, and then normalized to zero mean and unit variance to compensate for different units of measure. Finally, data were re-sampled from 30 Hz to 5 Hz for shorter computation time.

In order to learn the task progress, new ground truth labels were automatically generated from the available data and action labels. 
As a preliminary step, however, we carefully inspected the video recordings in order to identify possible imprecisions in the available annotations, that would affect the automatic generation of our progress labels. We identified and corrected 12 mistakes, affecting a total of 2356 data samples. 
Amendments to the original annotations are reported in the Appendix.

As illustrated in Fig. \ref{fig:progLabels}, our definition of progress is dictated by the underlying action sequence. 
Out of the 10 original action labels from JIGSAWS, we identified five gestures that constitute essential progressive stages in any complete suturing demonstration (\textit{Reaching for the needle, Positioning the tip of the needle, Pushing the needle through the tissue, Pulling the suture, Dropping the suture}), generating a simplified probabilistic state machine that describes the commonly-observed workflow of the suturing task. 
The other classes represent adjustment gestures that serve to prepare or help to complete the execution of the essential gestures and that generally appear in variable order.
We thus grouped fundamental gestures (performed by any of the two arms, even if JIGSAWS only features right-handed suturing demonstrations) and their corresponding adjustment gestures into 5 progress stages (from 0 to 4), as detailed below:
\begin{itemize}  
	\item Progress 0: G1 \textit{Reaching for needle with right} + G5 \textit{Moving to center of workspace}
	\item Progress 1: G2 \textit{Positioning the tip of the needle} + G4 \textit{Transferring the needle from left to right} before G2 + G8 \textit{Orienting the needle} before G2
	\item Progress 2: G3 \textit{Pushing the needle through the tissue} + G4/G8 before G3
	\item Progress 3: G6 \textit{Pulling the suture with left} + G9 \textit{Using right hand to tighten suture} + G10 \textit{Loosening more suture} + G4/G8 before G6/G9/G10
	\item Progress 4: G11 \textit{Dropping suture and moving to end points} + G4/G8 before G11
\end{itemize}
As the task evolution in time is affected by numerous factors, such as surgical skill and surgical context, we believe that activity-based progress could be better than time-based progress in reducing the kinematic feature variation for equal progress values. Moreover, it could help to learn action sequentiality despite the presence of adjustment gestures which occur in variable frequency and uncertain order.\\ 

\begin{figure*}[t]
	\centering
	\includegraphics[width=0.93\textwidth]{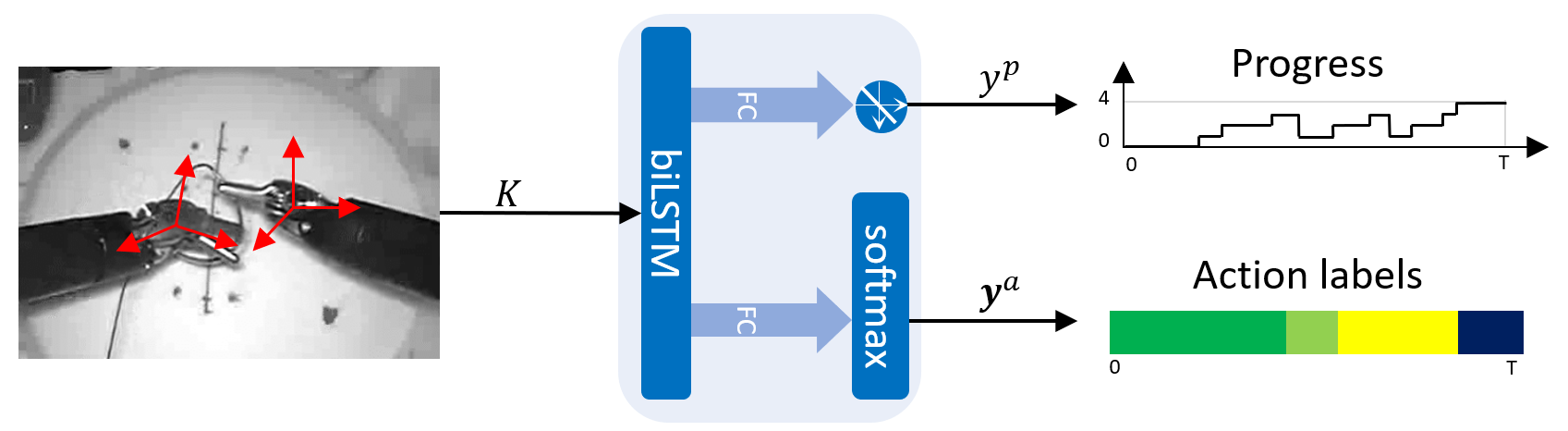}
	\caption{Multi-task architecture for joint action recognition and progress regression. The kinematic features (K) are fed to a bidirectional LSTM cell, whose hidden units are connected to the regression node by a fully connected layer. The same hidden units are projected by a second fully connected layer into 10 logits with softmax activation function for action classification.}
	\label{fig:Architecture}
\end{figure*}

\subsection{Multi-task Recurrent Neural Network}

Our multi-task architecture performs action recognition jointly with progress estimation. As the progress is quantized into 5 sequential steps, we estimate it using three different strategies: regression, standard classification and classification with ordered classes (or ordinal regression).

Notation: vectors are represented in bold lowercase letters (e.g. \textbf{y}), scalars in lowercase letters (e.g. $y$), parameters and losses in uppercase letters (e.g. C).\\

\begin{figure}[t]
	\centering
	\includegraphics[width=\columnwidth]{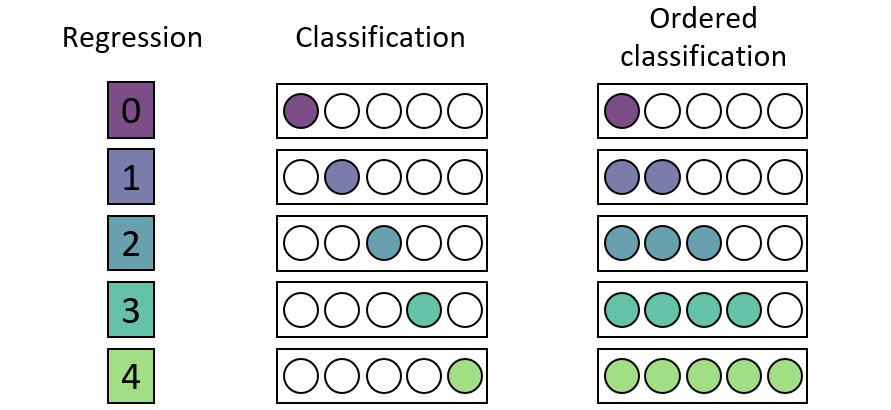}
	\caption{Target encodings for progress regression, classification and ordered classification.}
	\label{fig:Targets}
\end{figure}

\subsubsection*{\textit{Regression}}
As shown in Fig. \ref{fig:Architecture}, the kinematic features ($K$) are fed to a single-layer bidirectional LSTM with 1024 hidden units. Activations from the forward and backward streams are concatenated into a 2048-dimensional vector and then connected to the regression node by a Fully Connected (FC) layer with linear activation function. 
The same 2048 features are also projected by a second fully connected layer into 10 logits with softmax activation function for action classification.

At each training iteration, we compute the regression loss using the Mean Absolute Error (MAE) over individual demonstrations:
\begin{equation}
	MAE=\frac{1}{T}\sum_{t=1}^{T}{|y^{p}_{t}-\hat{y}^{p}_{t}|},
\end{equation}
and the classification loss using the Mean Cross Entropy (MCE) over individual demonstrations, as in \cite{DiPietro2016}:
\begin{equation}
	MCE=\frac{1}{T}\sum_{t=1}^{T}{\bigg(-\sum_{c=1}^{C}{y_{tc}^{a}\log(\hat{y}_{tc}^{a})}\bigg)},
\end{equation}
where $T$ is the demonstration length (number of samples), $C$ is the number of action classes, $y^{p}_{t}$ and $\textbf{y}^{a}_{t}$ are the regression and prediction nodes' output at timestamp t, and $\hat{y}^{p}_{t}$ and $\hat{\textbf{y}}^{a}_{t}$ are the corresponding ground truths.

After model training, the regression output is rounded to the nearest integer for progress prediction, and the logit with largest activation is considered for action prediction.\\

\begin{figure*}[t]
	\vspace{0.05 cm}
	\centering
	\includegraphics[width=0.85\textwidth]{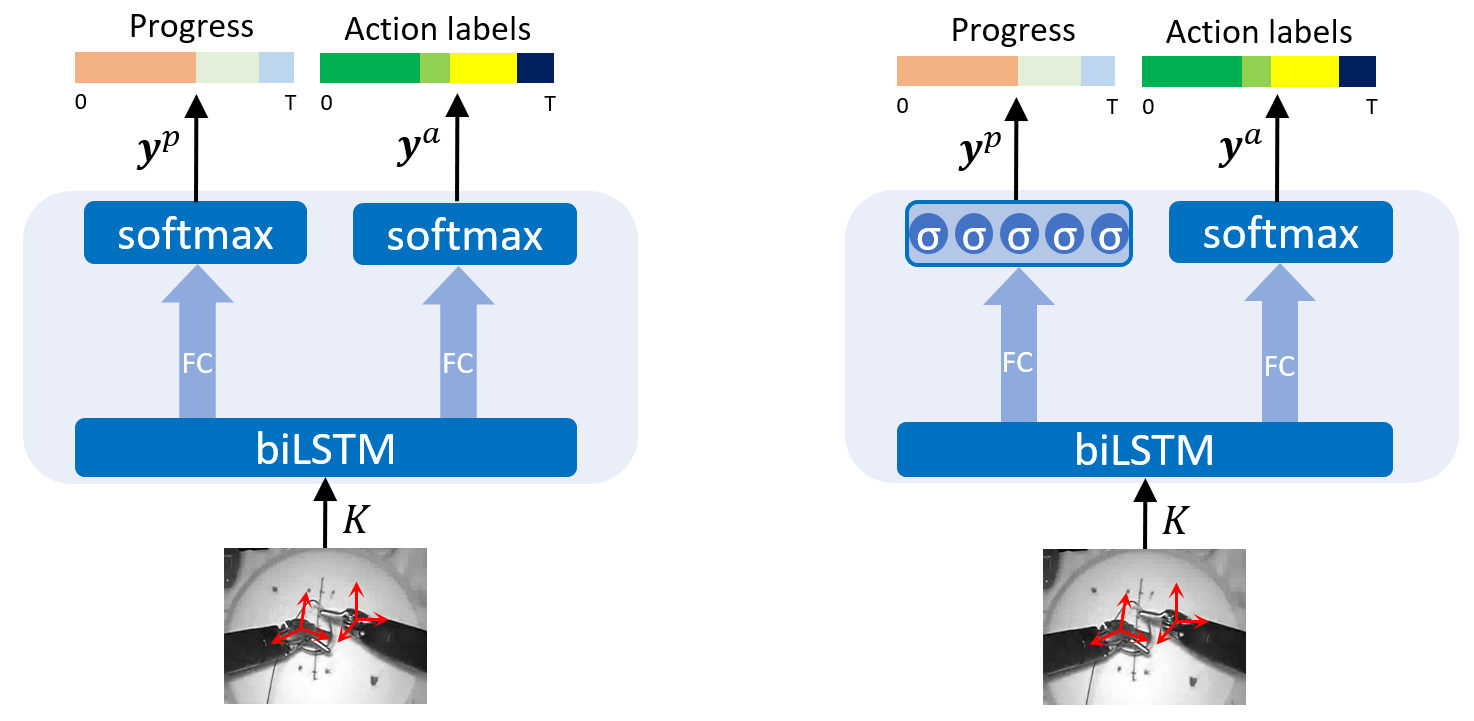}
	\caption{On the left: multi-task architecture for joint action recognition and progress classification. The LSTM hidden units are connected to 5 logits with softmax activation and MCE loss for progress classification. On the right: multi-task architecture for joint action recognition and ordered progress classification. The LSTM hidden units are connected to 5 logits with sigmoid activation and MBCE loss for ordered progress classification.}
	\label{fig:Architecture2}
\end{figure*}

\subsubsection*{\textit{Classification}}

To perform standard progress classification, we substitute the regression layer with a 5-logit fully connected layer with softmax activation function and MCE loss, thus obtaining a multi-hierarchical action recognition network (Fig. \ref{fig:Architecture}). After model training, the logit with largest activation is considered for progress prediction.\\ 

\subsubsection*{\textit{Ordered classification}}

Standard classification considers independent categories and does not penalize major ordering mistakes. 
In order to represent the succession of progress classes, we thus encoded the target vectors with the ordinal formulation of \cite{Cheng2008} as represented in Fig. \ref{fig:Targets}, and substituted the categorical MCE loss with the Mean Binary Cross Entropy (MBCE) loss (i.e. Sigmoid activation function and MCE loss). MBCE sets up an independent binary classifier for each class and, in combination with the ordinal target encoding, generates a larger loss the further the prediction is from its ground truth. After model training, progress predictions are obtained from the output $\textbf{y}^{p}$ of this classifier by finding the first index $k$ where $y_{k}^{p}$\textless 0.5.\\

In all the three cases, the final multi-task loss ($L$) is a weighted combination of the two single-task losses ($L_{1}=MCE$, $L_{2}=MAE\ or\ MCE\ or\ MBCE$):
\begin{equation}
L=w_{1}L_{1}+w_{2}L_{2} 
\end{equation}

However, multi-task networks are generally difficult to train, as task imbalances may lead to the generation of shared features that are not useful across all tasks. In order to automatically balance our model training, we used the GradNorm algorithm \cite{Chen2017} for gradient normalization, that has been shown to improve accuracy and reduce overfitting across multiple tasks when compared to single-task networks. GradNorm dynamically updates the single-task loss weights ($w_{1}$, $w_{2}$) during training by optimizing an additional loss ($Lgrad$), which aims at regularizing the training rate of the individual tasks:
\begin{equation}
Lgrad=\sum_{i=1}^{2}{\left|g_{\textbf{w}}^{(i)}-\bar{g}_{\textbf{w}}\times(r_{i})^{\alpha}\right|}
\end{equation}

$g_{\textbf{w}}^{(i)}=||\bigtriangledown_{\textbf{w}} w_{i}L_{i}||$ is the $L_{2}$-norm of the gradient of the weighted single-task loss $w_{i}L_{i}$ with respect to the network weights $\textbf{w}$.

$\bar{g}_{\textbf{w}}$ is the average gradient norm across all tasks.

$r_{i}=(L_{i}/L_{i}^{0})/\bar{L}$ is the relative inverse training rate of task $i$, with $L_{i}^{0}$ the single-task loss at the first training iteration and $\bar{L}$ the average loss across all tasks.

$\alpha$ is a balancing hyperparameter to be tuned.\\

\subsection{Evaluation Setup}

As in \cite{Lea2017tcnCVPR}, we evaluated our network recognition performance using accuracy, i.e. the percentage of correctly labelled frames, normalized segmental Edit score, which determines the precision of the predicted temporal ordering of actions, and segmental F1@10 score, which penalizes over-segmentation errors but is not sensitive to minor temporal shifts between predictions and ground truth. Progress regression was evaluated with MAE, normalized with respect to the full range of progress values ($\frac{MAE}{4-0}*100$). 

We followed the standard JIGSAWS Leave One User Out (LOUO) cross-validation setup \cite{Gao2014}: for every fold, all the trials performed by a single user are kept out as the test set and the other demonstrations are used to train our model.

\begin{table*}[t]
	\vspace{0.5cm}
	\caption{Gesture recognition (A) and progress estimation (P) performance. Scores are represented as mean(std).}\label{tab1}
	\centering
	\begin{tabular}{cc|c|cc|cc|cc|c|c|c}
		\hline
		\multicolumn{2}{c|}{Scores} & {\bfseries A\T\B} & \multicolumn{2}{c|}{\bfseries APr\T\B} &  \multicolumn{2}{c|}{\bfseries APc\T\B} &  \multicolumn{2}{c|}{\bfseries APoc\T\B}& {\bfseries Pr} & {\bfseries Pc} & {\bfseries Poc}\\
		\T\B&   & 	& $w_{1}$=$w_{2}$=1 & $\alpha$=1.5 & {$w_{1}$=$w_{2}$=1} & $\alpha$=1.5 & $w_{1}$=$w_{2}$=1 & $\alpha$=1.5 &  & & \\
		\hline
		\multirow{3}{0.5em}{A} & {\bfseries Accuracy\T} &85.3(5.8) &85.1(6.6)  &85.2(6.3) &85.7(5.7) &85.8(5.6) &85.5(5.8) &\textbf{86.0(5.4)} &- &- &- \\
		& {\bfseries Edit}         					&83.1(7.4) &84.2(6.7)  &85.9(6.4) &85.4(5.7) &85.7(5.7) &\textbf{86.2(6.3)} &86.1(6.2) &- &- &-\\
		& {\bfseries F1@10\B} 			 					&88.5(5.7) &89.0(5.5)  &90.1(5.4) &90.1(4.9) &90.2(4.9) &90.5(5.0) &\textbf{90.7(4.8)} &- &- &-\\
		\hline
		\multirow{4}{0.5em}{P} & {\bfseries MAE\T}     &-  	   &5.3(1.5)  &\textbf{5.1(1.5)}   &- 		 &- 		&- 		   &-         &6.2(1.1) &- &-\\
		& {\bfseries Accuracy}           					&-         &-  	      &-          &89.0(2.7) &89.1(2.8) &87.9(3.5) &88.4(3.1) &- &\textbf{89.2(2.8)} &87.0(4.0)\\
		& {\bfseries Edit}         					&-  	   &-  	      &-          &\textbf{89.3(6.7)} &89.2(6.6) &83.7(4.8) &83.6(4.5) &- &87.8(7.8) &87.3(5.7)\\
		& {\bfseries F1@10\B}            					&- 		   &-  	      &-          &93.2(4.4) &\textbf{93.2(3.4)} &90.0(3.1) &90.0(3.1) &- &91.9(4.9) &91.9(3.6)\\
		
		\hline
	\end{tabular}
	\label{table:Results}
\end{table*}

\section{Experiments and Results}

We used the open source TensorFlow implementation of the Bidirectional LSTM presented in \cite{DiPietro2016} as our baseline (A). We also relied on the provided training parameters, since they were carefully tuned on the same dataset. Given the stochastic nature of the optimization process, all experiments were performed three times and results were averaged. All runs were trained on NVIDIA Tesla V100-DGXS GPU, with training time of about 1 hour per run.

Our multi-task network for joint action recognition and progress regression (APr) was learnt on the multi-task loss $L$ using Gradient Descent (GD) with Momentum 0.9, batch size 5 and initial learning rate 0.1. The multi-task architectures for standard progress classification (APc) and ordered progress classification (APoc) were instead trained with GD, batch size 5 and initial learning rate 1.0. We always applied learning rate decay of 0.5 after 80 iterations and stopped the training after 120 iterations.
We used gradient clipping to avoid exploding gradients and dropout regularization with dropout rate of 0.5, as for the baseline (A). The single-task loss weights ($w_{1}$, $w_{2}$) were updated at a learning rate of 0.025 using GD on the regularization loss ($Lgrad$), with $\alpha$ set to 1.5.
Testing was performed after 100, 110 and 120 training iterations and results were averaged.
We trained all networks on the pre-processed kinematic data with revised annotations.\\

Comparison between A, APr, APc and APoc is presented in Table \ref{table:Results}. Multi-task performance is evaluated with ($\alpha=1.5$) and without ($w_{1}$=$w_{2}$=1) GradNorm regularization. Scores are reported as mean values across the 8 validation folds and corresponding standard deviations, which are strongly representative of inter-surgeon style variability in the LOUO setup. All three multi-task architectures outperform the single-task baseline on the segmental scores (Edit and F1@10), which seems to confirm the hypothesis that action-based progress estimation could help to learn action sequentiality and to reduce out-of-order predictions and over-segmentation errors. Even if none of the proposed architectures clearly stands out from the others, APoc generates slightly better results, which could be explained by stronger penalization of major ordering mistakes than standard classification, and easier optimization goal than regression of a discontinuous progress function. The architecture that benefits the most from multi-task gradient normalization is APr, as it is perhaps more challenging to balance two different loss functions (MCE for classification and MSE for regression) than two similar or identical ones. 
However, balanced multi-task networks rely on a large number of hyperparameters, including optimization parameters for the regularization loss. We believe that results could be improved and differences between the three proposed architectures could be emphasized with more extensive parameter tuning, as well as with larger datasets.

\begin{figure}[t]
	\centering
	\includegraphics[width=0.98\columnwidth]{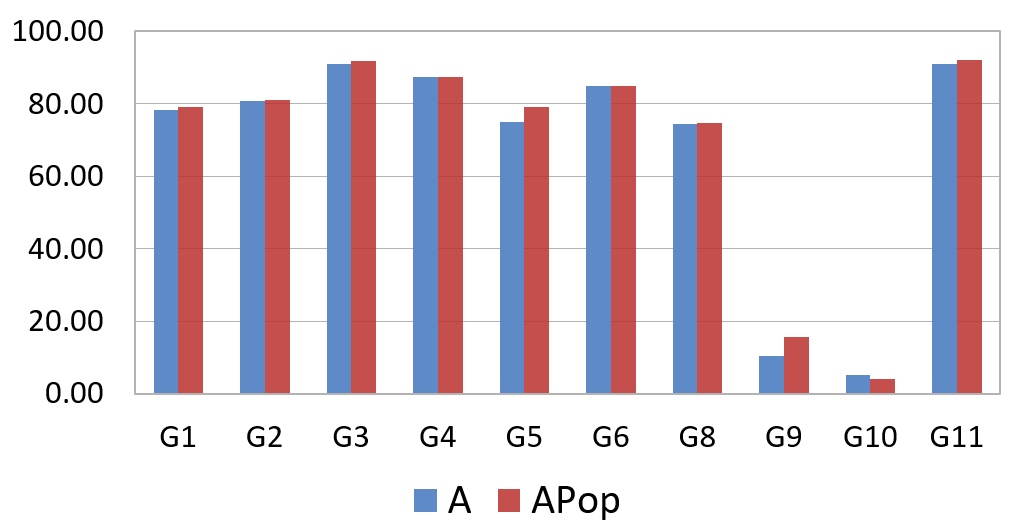}
	\caption{Recognition accuracy [\%] of individual gestures.}
	\label{fig:gestures}
\end{figure}

Fig. \ref{fig:gestures} shows recognition accuracies of individual gestures from A and APoc. APoc consistently matches or outperforms A, even if improvement is only marginal. Results in Table \ref{table:Results}, however, showed that the advantage of the proposed method relies in the regularization of the predicted sequences, which mainly affects the segmental scores and only marginally the framewise evaluation metrics. Some gestures, such as G9 and G10, are extremely challenging to recognize in both cases, as they are under-represented in the dataset.

In addition to recognising surgical gestures, our multi-task architectures segment the surgical demonstrations into 5 fundamental progressive steps of the suturing task, reaching an average accuracy of 89.1\% with standard classification (Table \ref{table:Results}). For APr and APc, but not for APoc, all evaluation scores improved with respect to their single-task counterparts Pr and Pc\footnote{Pr, Pc and Poc were trained once with the same hyperparameters as their multi-task counterpart. Weight decay was however anticipated and training was stopped after 80 iterations.}: not only higher-level progress understanding can help gesture recognition, but gesture recognition can reciprocally boost progress prediction. 

\begin{figure}[t]
	\vspace{0.25 cm}
	\centering
	\includegraphics[width=0.98\columnwidth]{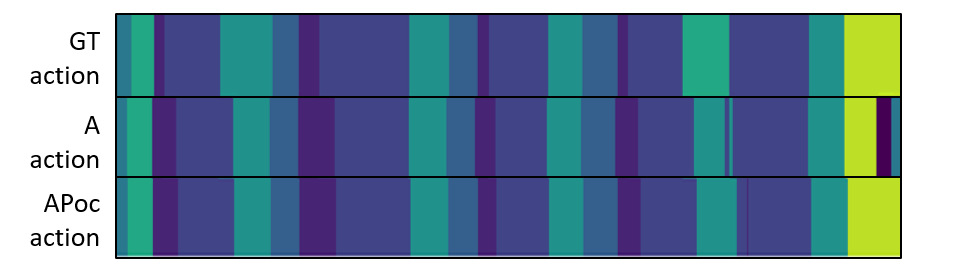}
	\caption{Example of recognition output where predictions generated by the  multi-task network (APoc) show reduced over-segmentation with respect to the baseline (A), as quantified by improved segmental scores. The ground truth segmentation (GT) is shown at the top.}
	\label{fig:results}
\end{figure}

\begin{figure}[t]
	\vspace{0.15 cm}
	\centering
	\includegraphics[width=1.03\columnwidth]{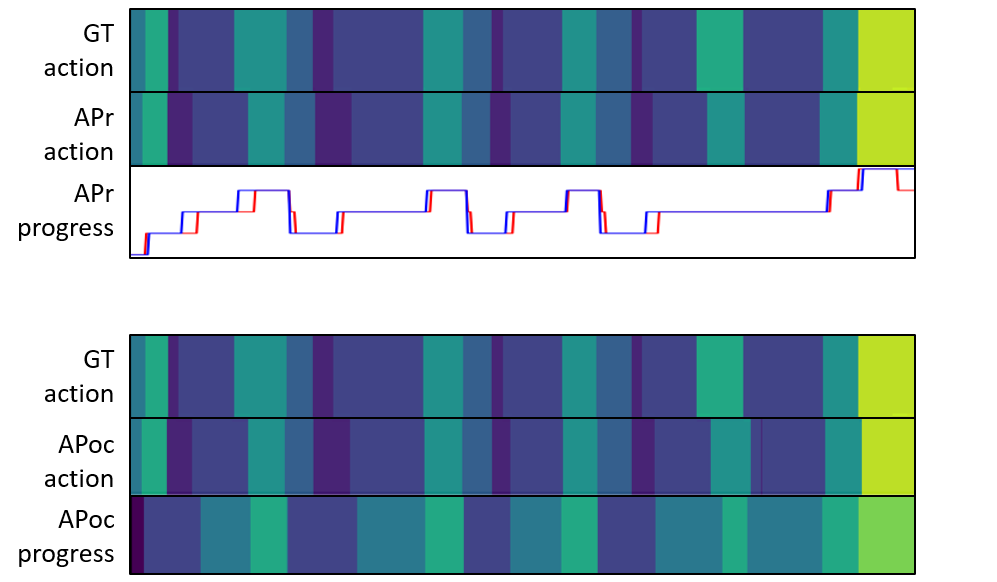}
	\caption{Comparison between action and progress predictions (APr progress prediction in red, progress ground truth in blue). The predicted segmentation boundaries are frequently aligned.} 
	\label{fig:results2}
\end{figure}

\begin{figure}[t]

	\vspace{0.2 cm}
	\centering
	\includegraphics[width=0.9\columnwidth]{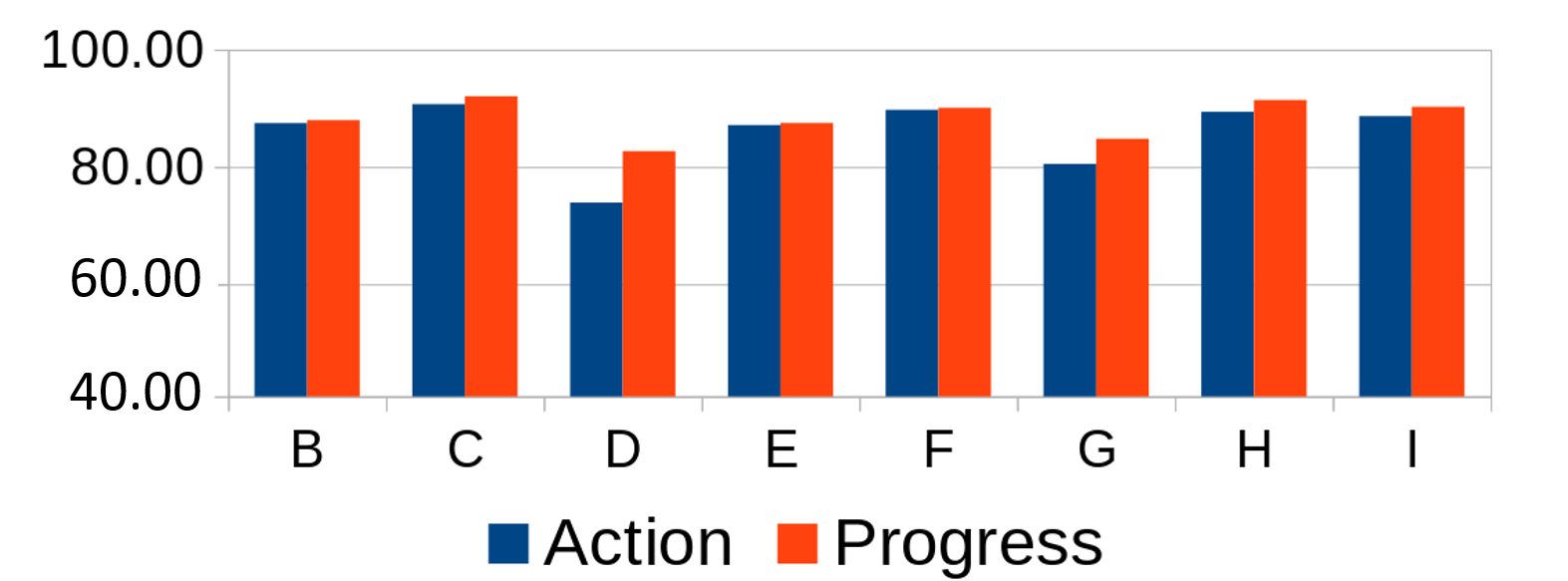}
	\caption{Action and progress prediction accuracy [\%] from APoc for each cross-validation fold (B, C, D, E, F, G, H, I). Poor progress estimation often corresponds to poor gesture recognition, and vice versa.}
	\label{fig:correl}
\end{figure}

Fig. \ref{fig:results} illustrates an example of recognition output where predictions generated by the  multi-task network show reduced over-segmentation with respect to the baseline, as quantified by the segmental score improvement previously reported. It is also interesting to visualize the relationship between gesture and progress predictions, as the segmentations boundaries are frequently aligned (Fig. \ref{fig:results2}), and poor progress estimation often corresponds to poor gesture recognition, and vice versa (Fig. \ref{fig:correl}).

\begin{table}[t]
	\caption{Comparison with related work on robot kinematics.}\label{tab1}
	\centering
	\begin{tabular}{c|c|c}
		& {\bfseries Accuracy\T\B} &  {\bfseries \ \ \ Edit \ \ \ \T\B} \\
		\hline
		{\textbf{TCN}\cite{Lea2016tcnECCV}\T} &79.6 &85.8 \\
		{\textbf{TCN+RL}\cite{Liu2018}} &82.1 &\textbf{87.9} \\
		{\textbf{BiLSTM}\cite{DiPietro2016}} &83.3 &81.1 \\
		{\textbf{APoc}} &85.3 &84.5 \\
		{\textbf{APc}\B} &\textbf{85.5} &85.3 \\
		\hline
	\end{tabular}
	\label{table:relatedwork}
\end{table}

We also trained APc and APoc with the original annotations of JIGSAWS, in order to compare our multi-task models to the original single-task baseline \cite{DiPietro2016} and to related work on robot kinematics. 
Our investigation, however, was carried out on a simple LSTM architecture, and we suggest the proposed multi-task approach could be applied on top of more complex architectures to boost performance. Results in Table \ref{table:relatedwork} highlight sensitivity of our models to action annotation noise, which partially spoils the automatic generation of progress labels. This results in performance degradation with respect to the previous experiments, especially for APoc. Nonetheless, the proposed networks significantly outperform 
\cite{DiPietro2016} both in accuracy and Edit score, 
and reach competitive performance with respect to related work on robot kinematics.

Finally, we substituted the Bidirectional LSTM cell in APc with a Forward LSTM cell for online recognition. We reached accuracy and Edit score of 82.2 and 76.2 respectively, improving upon the original single-task baseline \cite{DiPietro2016} (Table \ref{table:online}).

Our results support the hypothesis that 
joint surgical gesture recognition and progress estimation can induce more robust feature learning than gesture recognition alone, and boost performance in both online and offline applications.

\begin{table}[t]
	\vspace{0.15cm}
	\caption{Online gesture recognition performance.}\label{tab1}
	\centering
	\begin{tabular}{c|c|c}
		& {\bfseries Accuracy\T\B} &  {\bfseries \ \ \ Edit \ \ \ \T\B} \\
		\hline
		{\textbf{LSTM}\cite{DiPietro2016}\T} &80.5 &75.3 \\
		{\textbf{APc}} &\textbf{82.2} &\textbf{76.2} \\
		\hline
	\end{tabular}
	\label{table:online}
\end{table}

\section{Conclusions}

In this paper, we performed joint recognition of surgical gestures and progress prediction from robot kinematic data. Differently from prior work, the progress labels were defined on the underlying action sequence rather than on time, in order to reduce kinematic feature variation for equal progress values. Moreover, adjustment gestures did not contribute to the progress advancement. We assumed that action-based progress prediction could help to recognize surgical gestures in well-structured tasks such as suturing and knot tying, which are generally performed several times during surgical interventions.
We analysed different progress estimation strategies, and demonstrated on the suturing demonstrations of the JIGSAWS dataset that the proposed multi-task networks outperform the single-task baseline in terms of Edit score and F1@10 score, indicating a reduction in out-of-order predictions and over-segmentation errors. 
Since action-based progress does not depend on time nor on adjustment gestures, we conjecture this approach could also be effective beyond JIGSAWS in unconstrained environments, such as real surgical interventions or free surgical training sessions, where demonstrations do not have standardized length, right and left hands are often used interchangeably, and adjustment gestures, pauses and undefined motions are more frequent. 
In this scenario, contextualization of surgical motion into high-level progress stages could help to better recognize the surgical actions. 
The limitation of this method, however, is in the recognition of unstructured tasks such as blunt dissection, where action-based progress can not be clearly defined. 
In the presence of frequent and scattered mid-task failures and restarting, the ordered classification method might also lose its advantage to the standard classification method.

As suggested in \cite{Yengera2018}, further investigation could be performed on alternative multi-task integration modalities, such as pre-training on the auxiliary task for feature extraction or fine-tuning on the target task. This might potentially match or even improve upon multi-task training, at the cost of additional training time. 
Another study could model the progress in time of the individual gestures, which could improve understanding of gesture evolution and duration. 
Moreover, the integration of visual features extracted from surgical videos could boost both action recognition and progress estimation, as video data encode complementary information about the surgical tools and the state of the environment. 

Finally, evaluation of the proposed methodology was performed on the JIGSAWS dataset, which is currently the only publicly available dataset for surgical gesture recognition featuring robot kinematics. However, JIGSAWS is small and contains a limited range of surgical motions.
New surgical data will be collected in the future, and extensive evaluation will be carried out on larger datasets of robotic surgical demonstrations.





\section*{APPENDIX}

Amendments to the original annotations of JIGSAWS:

\begin{table}[h]
	\centering
	\scriptsize
	\begin{tabular}{c|ccc}
		\textbf{Demonstration} &\textbf{Start} &\textbf{End} &\textbf{Label}\\
		\hline
		Suturing\_B004 \T &2650 &2860 &G3\\
		Suturing\_C002 &1596 &1685 &G4\\
		Suturing\_D003 &1013 &1250 &G9\\
		Suturing\_D003 &1251 &1339 &G4\\
		Suturing\_D004 &0099 &0166 &G5\\
		Suturing\_D004 &0167 &0275 &G8\\
		Suturing\_D004 &0956 &1020 &G4\\
		Suturing\_E003 &1095 &1267 &G4\\
		Suturing\_F001 &2401 &2498 &G6\\
		Suturing\_G001 &1132 &1353 &G6\\
		Suturing\_G001 &7628 &8181 &G8\\
		Suturing\_I003 &0800 &1250 &G3\\
		\hline
	\end{tabular}
\end{table}


%

\bibliographystyle{ieeetr}
\bibliography{library}

\end{document}